\documentclass[letterpaper, 10 pt, conference]{ieeeconf}  
\IEEEoverridecommandlockouts   

\usepackage{color,soul}
\usepackage{graphicx}
\usepackage[font=small,labelfont=bf,tableposition=top]{caption}

\usepackage{siunitx}
\usepackage{bbold}
\usepackage{hyperref}
\usepackage{subcaption}

\makeatletter
\let\NAT@parse\undefined
\makeatother
\usepackage[numbers]{natbib}

\newcommand{\revision}[1]{\textcolor{black}{#1}}

\newcommand{\underscoredw}[1]{\textit{#1}}

\title{
Interactive Gibson Benchmark (iGibson 0.5): \\
A Benchmark for Interactive Navigation in Cluttered Environments
}

\author{Fei Xia$^{1}$, William B. Shen$^{1}$, Chengshu Li$^{1}$, Priya Kasimbeg$^{1}$, Micael Edmond Tchapmi$^{1}$\\
Alexander Toshev$^{2}$, Li Fei-Fei$^{1}$, Roberto Mart\'in-Mart\'in$^{1}$, Silvio Savarese$^{1}$%
\thanks{$^{1}$ Stanford University. $^{2}$ Robotics at Google.}
\thanks{$^{3}$Project website: \url{https://sites.google.com/view/interactivegibsonenv}}
\thanks{$^{\dagger}$ The simulation environment proposed in this work is retrospectively named iGibson 0.5 to distinguish it from later versions.}
}

\begin{document}
\maketitle

\setcounter{footnote}{2}

\newcounter{fnnumber}

\vspace{-5mm}

\begin{abstract}
We present \textit{Interactive Gibson Benchmark}, the first comprehensive benchmark for training and evaluating \textit{Interactive Navigation} solutions. Interactive Navigation tasks are robot navigation problems where physical interaction with objects (e.g. pushing) is allowed and even encouraged to reach the goal. Our benchmark comprises two novel elements: 1) a new experimental simulated environment, the \textit{Interactive Gibson Environment (iGibson 0.5$^{\dagger}$)}, that generates photo-realistic images of indoor scenes and simulates realistic physical interactions of robots and common objects found in these scenes; 2) the \textit{Interactive Navigation Score}, a novel metric to study the interplay between navigation and physical interaction of Interactive Navigation solutions. We present and evaluate multiple learning-based baselines in Interactive Gibson Benchmark, and provide insights into regimes of navigation with different trade-offs between navigation, path efficiency and disturbance of surrounding objects. We make our benchmark publicly available$^3$ and encourage researchers from related robotics disciplines (e.g. planning, learning, control) to propose, evaluate, and compare their Interactive Navigation solutions in Interactive Gibson Benchmark.
\end{abstract}

\setcounter{fnnumber}{\thefootnote}%

\section{Introduction}
\label{s:intro}

\begin{figure}[t]
  \begin{center}
    \includegraphics[width=0.4\textwidth]{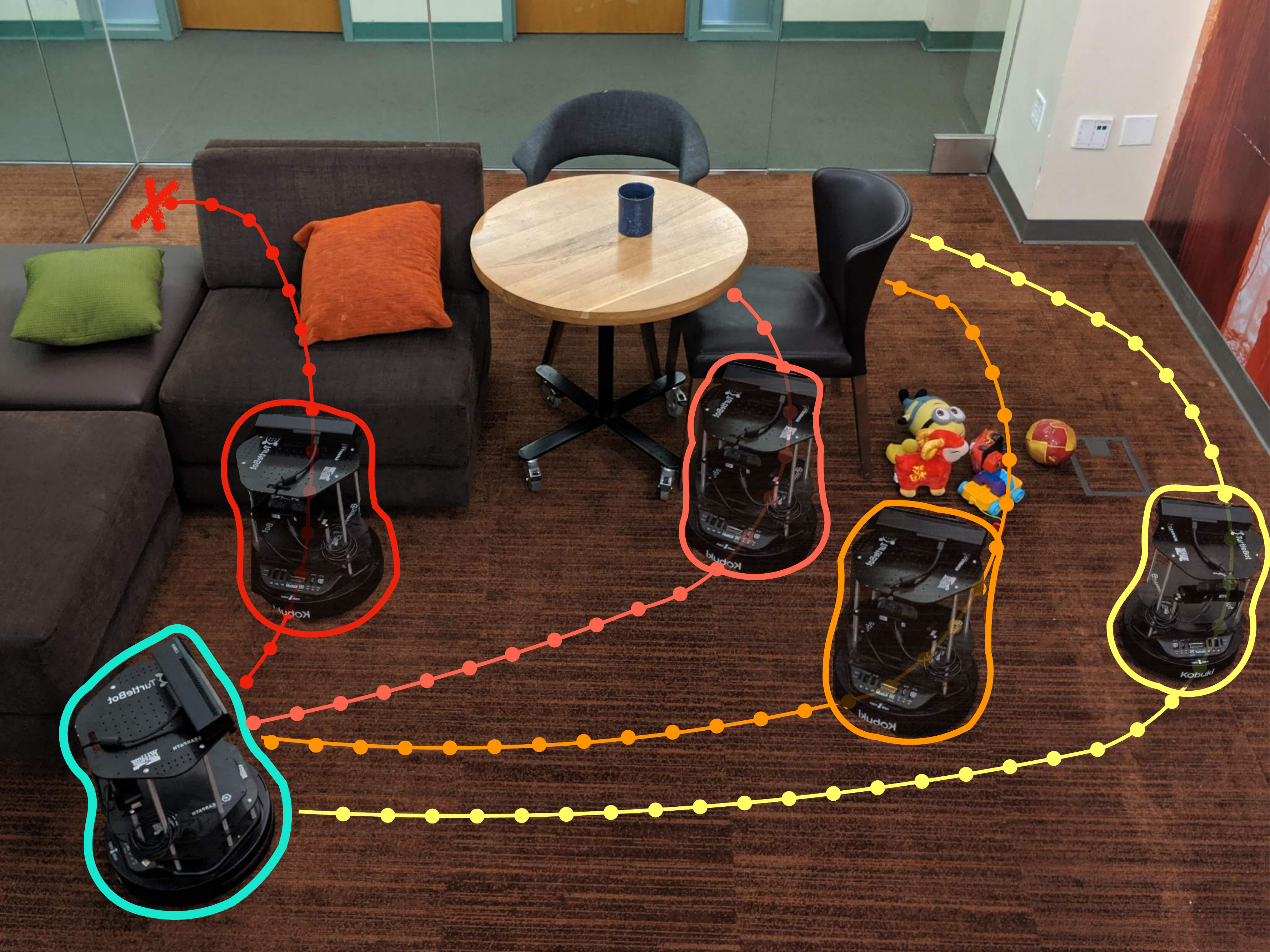}
  \end{center}
  \vspace{-2mm}
  \caption{\footnotesize We study \textbf{Interactive Navigation}, \revision{navigation} tasks where a mobile agent that has to navigate from its initial location (bottom-left, cyan) to a final goal (top-left, red cross) can use interactions with the environment as part of the strategy. We propose a simulator and a benchmark for this problem, called \textbf{Interactive Gibson Benchmark}, where we evaluate different modes of behavior, balancing optimality between path length (red,  shortest path length but unfeasible effort to interact) and effort due to interaction with the environment (yellow, no interaction but longest path).}
	\label{fig:pull}
	\vspace{-6mm}
\end{figure}

Classic robot navigation is concerned with reaching goals while avoiding collisions~\cite{Siciliano:2007:SHR:1209344,bonin2008visual}. This interpretation of navigation has been applied to a wide variety of robot navigation domains such as factories, \revision{warehouses} and outdoor settings. However, as robots are more often deployed in unstructured and cluttered environments \revision{such as homes and offices}, considering physical interaction part of the navigation strategy becomes not only unavoidable, but necessary. For example, when navigating in a cluttered home, a robot might need to push objects aside or open doors in order to reach its destination. This problem is referred to as \textit{Interactive Navigation} and in this paper we propose a principled and systematic way to study it (see Fig.~\ref{fig:pull}).


The ``aversion to interaction'' in robot navigation is easy to understand: real robots are expensive, and interacting with the environment creates safety risks. In robot manipulation, researchers have alleviated these risks using physics simulation engines~\cite{todorov2012mujoco,coumans2013bullet,koenig2004design}. Physics engines allow researchers to simulate object-robot dynamics with high precision in handcrafted models and to study manipulation in a safe manner. Researchers have further exploited these engines to train models that can be deployed in the real world. 

Unfortunately, the aforementioned simulation engines are not enough to study and train visual models for navigation: the handcrafted models lack the photorealism and complexity of the real-world spaces. As a result, in recent years we have seen a new class of simulation environments~\cite{kolve2017ai2,xia2018gibson,habitat19arxiv,chang2017matterport3d,juliani2018unity,armeni2017joint} that build upon renderers of \revision{synthetic or} real-world scans. These environments have a higher photo and layout realism, and provide realistic scene complexity. They facilitated the development and benchmarking of vision-based (learned) navigation algorithms, and some even enabled to deploy on real robots the algorithms developed in simulation~\cite{gupta2017cognitive,hirose2019deep}. Most of these simulators, however, fall short of interactivity -- they use realistic scanned models of real world scenes that are static, i.e., where objects cannot be manipulated. Such interactivity is crucial \revision{for navigation} in realistic cluttered environments, e.g., for Interactive Navigation problems. 

In this work, we study and present a benchmark for Interactive Navigation in cluttered indoor environments at scale. This benchmark is the result of multiple efforts: 1) providing a realistic simulation environment for the benchmark, 2) defining appropriate benchmark setup and metrics, and 3) providing and evaluating a set of (learning-based) solutions that act as baselines. Thus, the main contributions of this work are as follows.



First, we introduce a novel simulation environment, the \textit{Interactive Gibson Environment}, that retains the photo-realism and scale of the original Gibson Navigation environment~\cite{xia2018gibson} (Gibson V1 in the following) but allows for realistic physical interactions with objects. The new models in Interactive Gibson allow for the complex interactions between the agent and the environment \revision{such as pushing objects and opening doors}, which are part of Interactive Navigation solutions. Furthermore, compared to Gibson V1, Interactive Gibson Environment has significantly faster rendering speed making large-scale training possible. While in this work we only use Interactive Gibson Environment to study Interactive Navigation, the environment opens up new avenues to study and train novel visual navigation, manipulation and mobile manipulation solutions exploiting the synergies between navigation and interaction~\cite{li2019hrl4in}.


Second, we define the new \textit{Interactive Gibson Benchmark} in the Interactive Gibson Environment for Interactive Navigation agents. To evaluate solutions in the benchmark, we propose a novel performance metric, the \textit{Interactive Navigation Score} (INS), that unifies the two most relevant criteria: 1) the navigation success and path quality, and 2) the effort associated with the induced disturbance to the surroundings. The former is formalized via the recently proposed SPL metric~\cite{anderson2018evaluation} while the latter is proportional to the mass displaced and the force applied to the objects in the scene. Thus, this unifying metric captures a trade-off between path length and disturbance to the environment.

Finally, we present a set of learning-based baselines on two different robot platforms using established Reinforcement Learning (RL) algorithms such as PPO~\cite{schulman2017proximal}, DDPG~\cite{lillicrap2015continuous}, and SAC~\cite{haarnoja2018soft}. We show that by actively modulating the penalty for interactions in the reward function, we control the Interactive Navigation Score, and that different INS values correspond to different navigation behaviors in cases such as the one illustrated in Fig.~\ref{fig:pull}, where the agent can choose between navigating longer paths or pushing heavier objects. 

\section{Related Work}
\label{s:rw}

\textbf{Interactive Navigation:} While the literature on autonomous robot navigation is vast and prolific, researchers have paid less attention to navigation problems that require interactions with the environment, what we call Interactive Navigation. The robot control community have extensively studied in isolation the problem of opening doors with mobile manipulators~\cite{peterson2000high,schmid2008opening,petrovskaya2007probabilistic,jain2009behavior}. However, these approaches focus only on this interaction instead of studying an entire Interactive Navigation task.

Stilman {\it et al.}~\cite{stilman2005navigation} studied Interactive Navigation from a geometric motion planning perspective. In their problem setup, called Navigation Among Movable Objects (NAMO), the agent has to reason about the geometry and arrangement of obstacles to decide on a sequence of pushing/pulling actions to rearrange them to allow navigation~\cite{stilman2008planning,levihn13}.
However, their sampling-based planning solutions require knowledge of the geometry of all objects and environment, and the search problem in the NAMO setup is restricted to pushing/pulling in 2D space. We go beyond the NAMO setup into a realistic 3D physical problem where the agents are provided with partial observations of the environment.

\textbf{Evaluating and Benchmarking Robot Navigation:} In many scientific disciplines such as computer vision or natural language understanding, evaluation and benchmarking is achieved by curating a dataset and defining a set of evaluation metrics.
Due to robotics' real-world and interactive components, benchmarking is less straightforward, as one is to deal with different hardware and interactions with the environments~\cite{del2006benchmarks}.

As a result, the robot navigation community has proposed several formats for benchmarking. The most straightforward is to restrict benchmarking to the perceptual components related to robot navigation.
Geiger {\it et al.}~\cite{geiger2013vision} provides datasets and metrics to evaluate SLAM solutions and other localization skills under the assumption that these components are common to all navigation systems. This benchmark is becoming less relevant with the advent of novel navigation approaches that learn to map directly visual signals to navigation commands.

A different, more integrated approach to benchmark robot navigation is to provide experimental specifications to be reproduced by researchers~\cite{sprunk2016experimental}.
Despite the unquestionable realism and experimental reproducibility, the setup cost can be prohibitive. To overcome this barrier, Pickem {\it et al.}~\cite{pickem2017robotarium} provide remote access for users to run comparable experiments on the same physical robots. Due to the maintaining costs, this solution is not scalable to many users.

A different attempt to benchmark robot navigation is to organize competitions~\cite{ozguner2007systems, balch2002ten, braunl1999research, iagnemma2006editorial}. These have been organized as one-off events or on an annual basis. Despite the realism and fairness of such setups, their infrequency makes them less suitable for faster research development.

When it comes to robot navigation metrics, the overwhelming majority of algorithms are evaluated on navigation success -- getting successfully to the target. Failures are due to inability to find a path to target or collisions~\cite{nowak2010benchmarks}. A more complex set of metrics are concerned with various aspects of safety and path quality~\cite{munoz2007evaluation}. More precisely, safety is quantified by clearance from obstacles and traversal of narrow spaces. Quality is often quantified by path length with respect to the optimal path and smoothness~\cite{ceballos2010quantitative}. In simulation, the most recent benchmark for point-to-point navigation, HabitatAI~\cite{habitat19arxiv}, measures performance based only on path distance and success rate of reaching the goal~\cite{anderson2018evaluation}.

Previous metrics focused on safety and path quality are too restricting for Interactive Navigation problems. In Interactive Navigation tasks the agent has to manipulate objects to reach the goal, which would be labeled as collision by the above metrics. Further, none of the above metrics can be used to study the trade-off between the effort required to interact with objects or taking a longer path around them, a decision that agents in cluttered human environments often face.

\textbf{Robot Simulation Environments:} With the most recent improvements in realism of visual and physics simulation, as well as the increase and improvements of available assets and models, simulation engines are emerging as a scalable, realistic, and fair way to develop and evaluate (visual) navigation algorithms.


To generate virtual camera images, simulation environments \revision{typically} use either game engine renderers or mesh renderers. AI2-THOR~\cite{kolve2017ai2}, VRKitchen~\cite{gao2019vrkitchen}, and \revision{VRGym~\cite{xie2019vrgym}} are based on a game engine. The benefit of using a game engine is that it has a clean workflow to generate and incorporate game assets and design customized indoor spaces. The downside is that game engines are usually optimized for speed instead of physical fidelity, and the simulation of interactions is often tackled via \revision{scripting}. Furthermore, game engine renderers are proprietary, and the assets are expensive to acquire, \revision{difficult to generate, and requires manual design of the environments}. \revision{These constraints limit the scale and diversity of environments in simulators using game engines}.

A different class of simulation environments use 3D scanned models of real world environments. Examples are Gibson Environment~\cite{xia2018gibson} (V1), Habitat~\cite{habitat19arxiv}, and MINOS~\cite{savva2017minos}. The assets in these environments are generally of much larger number and scale, are photo-realistic and depict real world scene layouts. A downside is that there might be reconstruction artifacts due to the imperfection of 3D scanning, reconstruction, and \revision{texturing} technology. Further, since the 3D scans generate a single model, the assets are static environments that cannot be interacted. The Interactive Gibson Environment we present in this paper falls also under this category but goes beyond these environments by allowing to interact with the objects of the segmented classes in the scenes. Furthermore, our process of segmenting and replacing model parts with realistic-looking CAD models improves the quality of the models and reduces reconstructions artifacts. While not for navigation, it is worth mentioning a recent interactive environments specialized in the task of door opening~\cite{urakami2019doorgym}. The work we present here can be used similarly to train with multiple door models but we focus on full (interactive) navigation tasks.




\section{The Interactive Gibson Environment}
\label{s:method}


\label{ss:ige}


\begin{figure}[t]
  \begin{center}
    \includegraphics[width=0.4\textwidth]{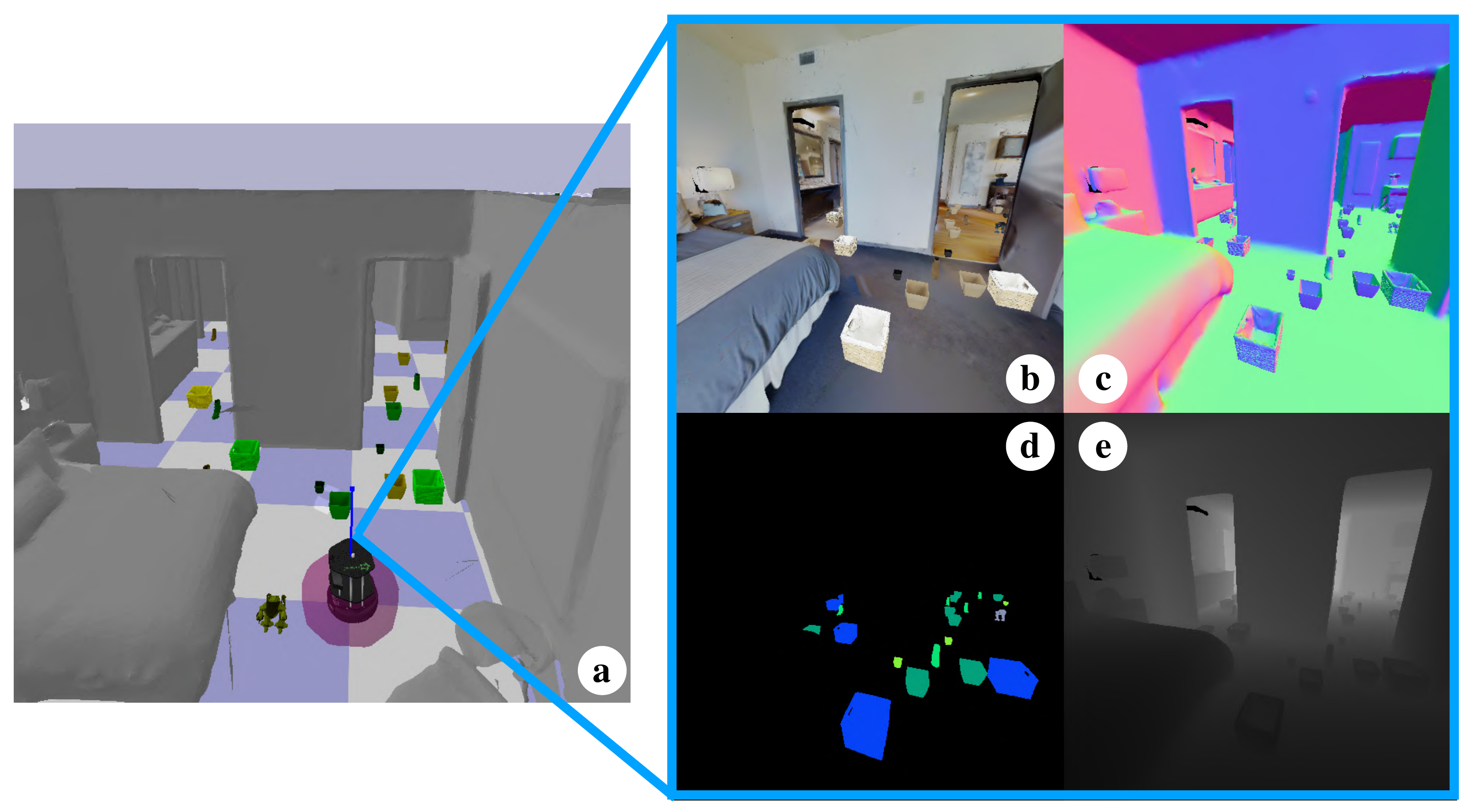}
  \end{center}
  \vspace{-2mm}
  	\caption{\footnotesize \textbf{Simulator and output modalities.} 3D view of the agent in the environment (a) and four of the visual streams provided by the Interactive Gibson Environment: RGB images (b), surface normals (c), semantic segmentation of interactable objects (d), and depth (e). In our experiments (Sec.~\ref{s:basalg}), only semantic segmentation and depth are used as inputs to our policy network.}
	\label{fig:simviews}
	\vspace{-6mm}
\end{figure}

The study of Interactive Navigation requires a reproducible and controllable environment where testing does not imply real risks for the robot. This advocates the use of simulation. Our previous work, Gibson V1~\cite{xia2018gibson}, provided a simulation environment to train embodied agents on visual navigation tasks without interactions. The main advantage of Gibson V1 is that it generates photo-realistic virtual images for the agent. This enabled seamless sim2real transfer~\cite{hirose2019deep,kang2019generalization}. However, Gibson V1 cannot be used as a test bed for Interactive Navigation because neither the rendering nor the assets (photo-realistic 3D models reconstructed from real-world) allow for changes in the state of the environment. 

We developed Interactive Gibson as a new simulation environment based on Gibson V1 with two main novelties. First, we present a new rendering engine that not only can render dynamical environments, but also runs extremely faster than Gibson V1's rendering, resulting in faster training of RL agents. Second, we present a new set of assets (scenes) where objects of relevant classes for Interactive Navigation (e.g. doors, chairs, tables, \ldots) can be interacted with.


\subsection{Interactive Gibson Renderer}

Gibson V1 performs {image-based rendering} (IBR)~\cite{shum2000review}. While achieving high photo-realism, IBR presents two main limitations. First, IBR is slow -- Gibson V1 renders at only 25-40 fps on modern GPUs. In order to render the scene, the system must load images from all available viewpoints and process them on-the-fly. This process is computationally expensive and limits the rendering \revision{speed} on most systems to close to real-time~\cite{hedman2016scalable}. For robot learning, especially sample inefficient methods such as model-free Reinforcement Learning~\cite{duan2016benchmarking}, IBR-based simulation can be prohibitively slow. 

Second, IBR can not be used for dynamic environments that change as result of interactions because these changes make the images taken from the initial environment configuration obsolete. Moving objects or adding new objects to the environment is thus not compatible with IBR, which impedes its usage for tasks like Interactive Navigation.

\begin{figure*}
\centering
  \includegraphics[width=0.9\textwidth]{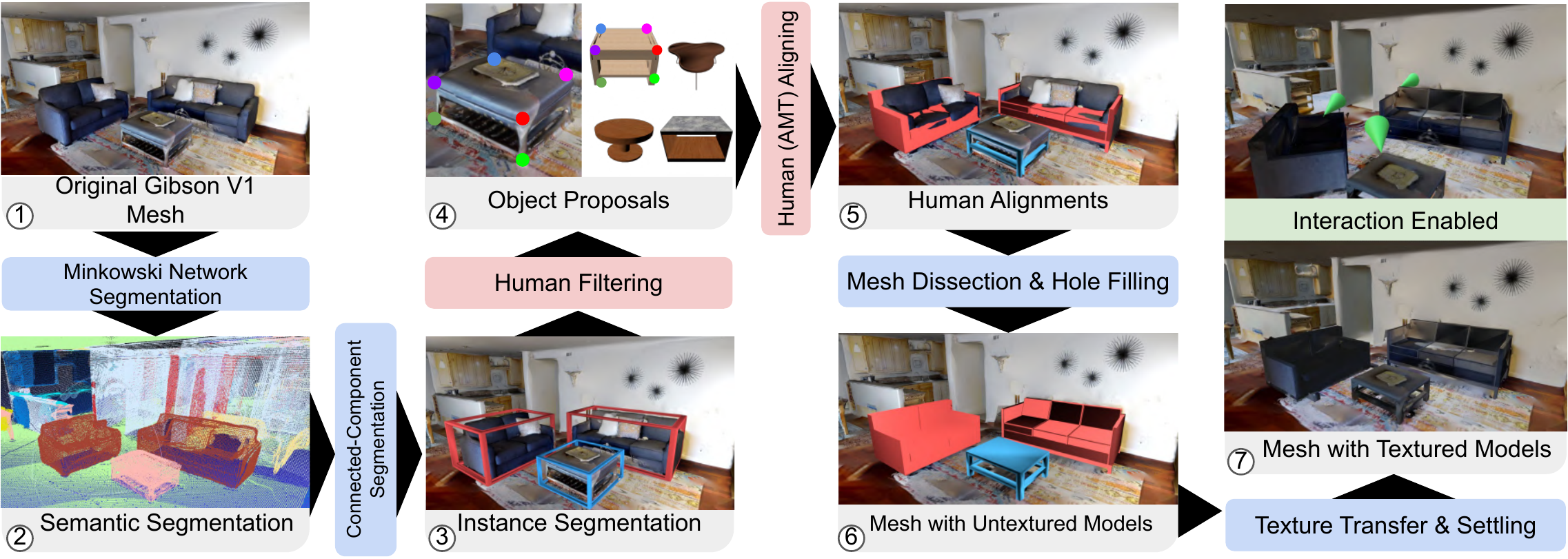}
  \caption{\footnotesize \textbf{Annotation Process of the Interactive Gibson Assets} In Gibson V1 each environment is composed by a single mesh (1); for Interactive Gibson we need to segment the instances of classes of \revision{interest} to study Interactive Navigation into separate interactable meshes; \revision{We use a combination of} a Minkowski SegNet~\cite{choy20194d} (2) and a connected component analysis (3) to generate object proposals (4). The proposals are manually aligned in Amazon Mechanical Turk (5) to the most similar ShapeNet~\cite{chang2015shapenet} model. Annotated objects are separated from the mesh and holes are filled (6), and the original texture is \revision{transferred} to the new object model (7) to obtain photo-consistent interactable objects, \revision{where one can apply forces as indicated by green cones}.}
  \label{fig:teaser}
  	\vspace{-6mm}
\end{figure*}

To overcome these limitations, in Interactive Gibson we replace image-based rendering with {mesh rendering}. This allows us to quickly train visual agents for Interactive Navigation tasks, where the agent not only navigates in the environment but also \revision{interacts} with objects.

Our high-speed renderer is compatible with modern deep learning frameworks because the entire pipeline is written in Python with PyOpenGL, PyOpenGL-accelerate, and pybind11 with our C++ code~\cite{pybind11}. This results in lower overhead, reduced computational burden, and a significant speedup of the rendering process (up to 1000 fps at $256\times256$ resolution in common computers).

To further optimize processes that rely on the results of the renderer (such as training vision-based RL agents), we enable a direct transfer of rendered images to tensors on GPUs. Avoiding downloading to host memory reduces device-host memory copies and significantly improves rendering speed.



\subsection{Interactive Gibson Assets}
\label{ss:iga}


Gibson V1~\cite{xia2018gibson} provides a massive dataset of high quality reconstructions of indoor scenes. However, each reconstruction consists of a single static mesh, which does not afford interaction or changes in the configuration of the environment (Fig.~\ref{fig:teaser}.1). For our Interactive Gibson Benchmark we augment the 106 scenes with 1984 interactable CAD model alignments of 5 different object categories: chairs, desks, doors, sofas, and tables. Our data annotation pipeline leverages both existing algorithms and human annotation to align CAD models to regions of the reconstructed mesh. To maintain the visual fidelity of replaced models, we transfer the texture from the original mesh to the CAD models.

Our assets annotation process (Fig.~\ref{fig:teaser}) is composed of the following combination of automatic and manual procedures (blue and pink blocks in Fig.~\ref{fig:teaser}): first, we automatically generate object region proposals using a state-of-the-art shape-based semantic segmentation approach (Fig.~\ref{fig:teaser}.2) with a further segmentation into instances (Fig.~\ref{fig:teaser}.3 and 4). These proposals are fed into a manual annotation tool~\cite{Avetisyan_2019_CVPR} where CAD models are aligned to the environment mesh. The resulting aligned CAD models (Fig.~\ref{fig:teaser}.5) are used to replace the corresponding segment of the mesh (Fig.~\ref{fig:teaser}.6) and the color of the original mesh is transferred to the CAD model to maintain visual consistency (Fig.~\ref{fig:teaser}.7) in the final interactable objects. Each stage of the pipeline is detailed below.

\textbf{Object Region Proposal Generation:}
Since Gibson V1 contains over 211,000 square meters of indoor space, it is infeasible to inspect the entire space by human annotators. We thus rely on an automated algorithm to generate coarse object proposals. These are areas of the reconstructed mesh of the environments that \revision{have} high probability of containing one or more objects of interest, and their corresponding class IDs. These proposals are then refined and further annotated by humans. We use a pretrained Minkowski indoor semantic segmentation model~\cite{choy20194d} to predict per-voxel semantic labels (Fig.~\ref{fig:teaser}.2). We then filter the semantic labels into instance segmentation (Fig.~\ref{fig:teaser}.3) through connected-component labeling~\cite{suzuki2003linear}. In areas with low reconstruction precision, the automatic instance segmentation results may contain duplicates as well as missing entries. These were manually corrected by in-house annotators (Fig.~\ref{fig:teaser}.4). In total, over 4,000 objects proposals resulted from this stage.

\textbf{Object Alignment:}
The goal of this stage is to 1) select the most similar CAD model from a set of possibilities~\cite{chang2015shapenet}, and 2) obtain the scale and the pose to align the CAD model to the reconstructed mesh. To obtain the alignments we use a modification of the Scan2CAD~\cite{Avetisyan_2019_CVPR} annotation tool. We crowdsourced object region proposals from the previous stage as HITs (Human Intelligence Tasks) \revision{on Amazon} Mechanical Turk\revision{ (AMT)} crowdsourcing market~\cite{doi:10.1177/1745691610393980}. 

The annotator is asked to retrieve the most similar CAD model from a list of possible shapes from ShapeNet~\cite{chang2015shapenet}. Then, the \revision{annotator} has to annotate at least 6 keypoint correspondences between the CAD model and the scan object (Fig.~\ref{fig:teaser}.4). The scale and pose alignment is solved by minimizing the point-to-point distance among correspondences over 7 parameters of a transformation matrix: scale (3), position (3), and rotation (1). Pitch and roll rotation parameters are predefined since the objects of interest almost always stand up-straight on the floor \revision{(in rest pose)}.

\textbf{Object Replacement and Re-texturing:}
Based on the alignment data, we process the corresponding region of the original mesh. We eliminate the vertices and triangular faces close to or inside the aligned CAD model. The resulting mesh contains discontinuities and holes. We fill them using a RANSAC~\cite{fischler1981random} plane fitting procedure (Fig.~\ref{fig:teaser}.6).

At this point we have replaced the parts of the reconstructed mesh by a CAD model. However, ShapeNet models are poorly textured. We improve visual fidelity and photo-realism by transferring the original texture from the images to the aligned CAD model~\cite{cignoni2008meshlab}. \revision{There will be small texture distortion cased by misalignment between scan and CAD model. But in practice the impact of texture distortion is not too large.} Finally, we correct for the small alignment noise in the CAD models' positions by running physics simulations to ensure they do not intersect with the floors and walls. For the dynamic properties of the objects that are relevant to interactions such as weight and friction, we assume a common set of parameters: density and material friction. Although this approximation may deviate from the real world values, we deem the effects of the deviation in the simulated interactions negligible. \revision{For articulated objects such as doors, we describe the structure of links and joints as a URDF (Unified Robot Description Format) model and we annotate it pose considering all joints in its origin (doors closed, drawers closed\ldots). For example, doors are composed by the frame and the leaf connected by a revolute joint (the hinge).}

The final result of the annotation is a new dataset of similar number of 3D reconstructed environments as the original Gibson V1 dataset, but where all objects of classes of interest for Interactive Navigation have been replaced by separate CAD models that can be interacted in simulation (Fig.~\ref{fig:teaser}.7).

\subsection{Interactive Gibson Agents}
\label{ss:igags}

Benchmarking Interactive Navigation requires embodied agents. We provide as part of Interactive Gibson ten fully functional robotic agents, including eight models of real robot platforms: two widely used simulation agents (the Mujoco~\cite{todorov2012mujoco} humanoid and ant), four wheeled navigation agents (Freight, JackRabbot v1, Husky and TurtleBot v2), a legged robot (Minitaur), two mobile manipulators with an arm (Fetch and JackRabbot v2), and a quadrocopter (Quadrotor). The large variety of embodiment types allows for easy tests of different navigation and interaction capabilities in Interactive Gibson.

The Interactive Gibson Environment enables a variety of measurements for the navigation agents (Fig.~\ref{fig:simviews}). The agents can receive as observations from the environment: 1) information about their own configuration such as position in the floor plan ({localization}), {velocity}, {collisions} with the environment and objects, motion ({odometry}), and visual signals that include photo-realistic {RGB images}, {semantic segmentation}, {surface normals}, and {depth maps}, and optionally 2) information about the navigation task such as {position of the goal}, and the ten closest next waypoints of the pre-computed {ground-truth shortest path} to the goal (separated by \SI{0.2}{\meter}). In Interactive Gibson Environment, the agents can control the position and velocity of each joint of their robot platform, including the wheels.

\section{Interactive Gibson Benchmark}
\label{ss:es}

The task of Interactive Gibson Benchmark is to navigate from a random starting location to a random goal location on the same floor. Both locations are uniformly sampled on the same floor place, and are at least \SI{1}{\meter} apart. 

\begin{figure}[t]
  \begin{center}
  \begin{subfigure}{\columnwidth}
         \centering
        \includegraphics[width=0.9\textwidth]{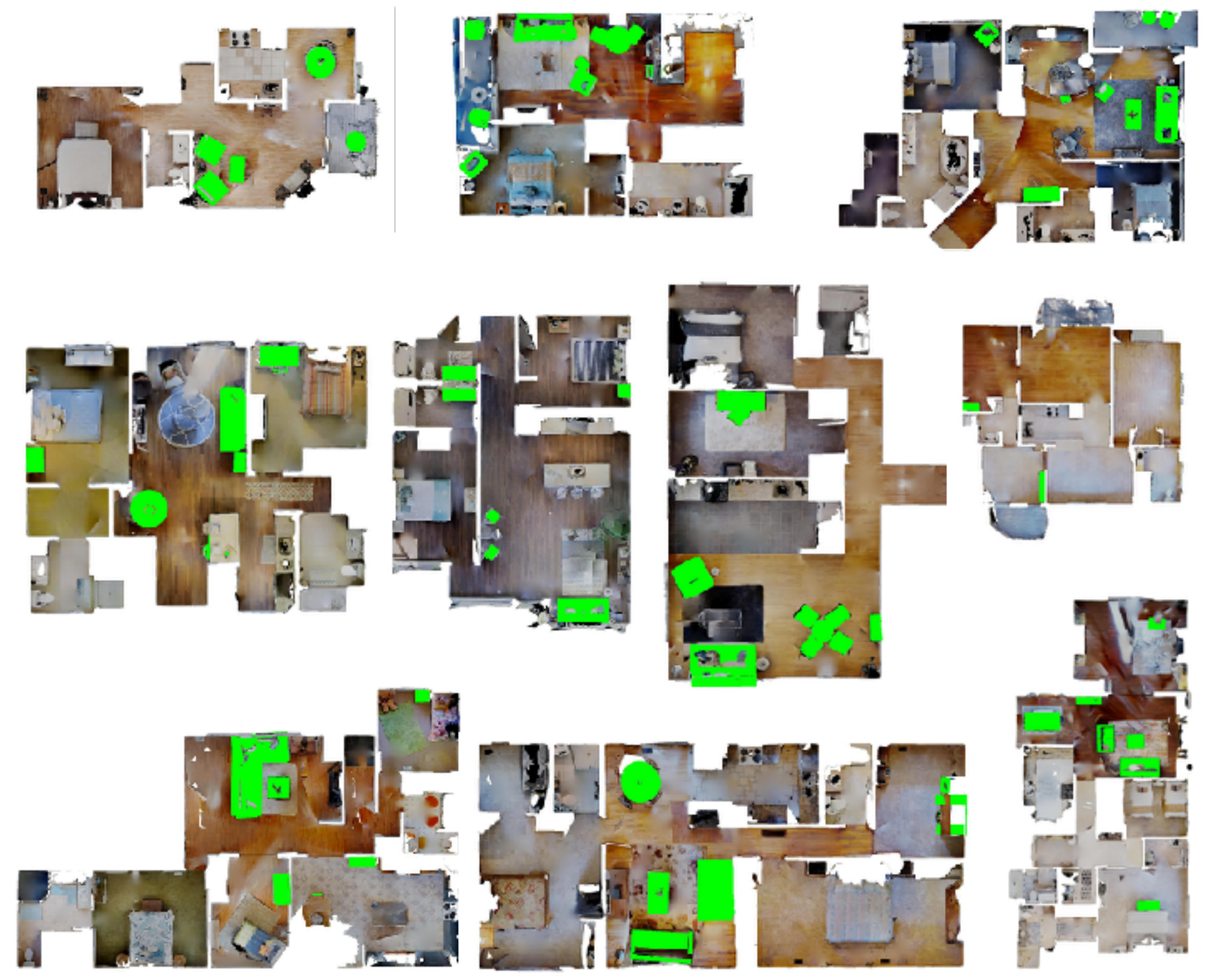}
        \caption{}
        \label{fig:added_objects:envs}
   \end{subfigure} 
\begin{subfigure}{0.4\columnwidth}
         \centering
        \includegraphics[width=\textwidth]{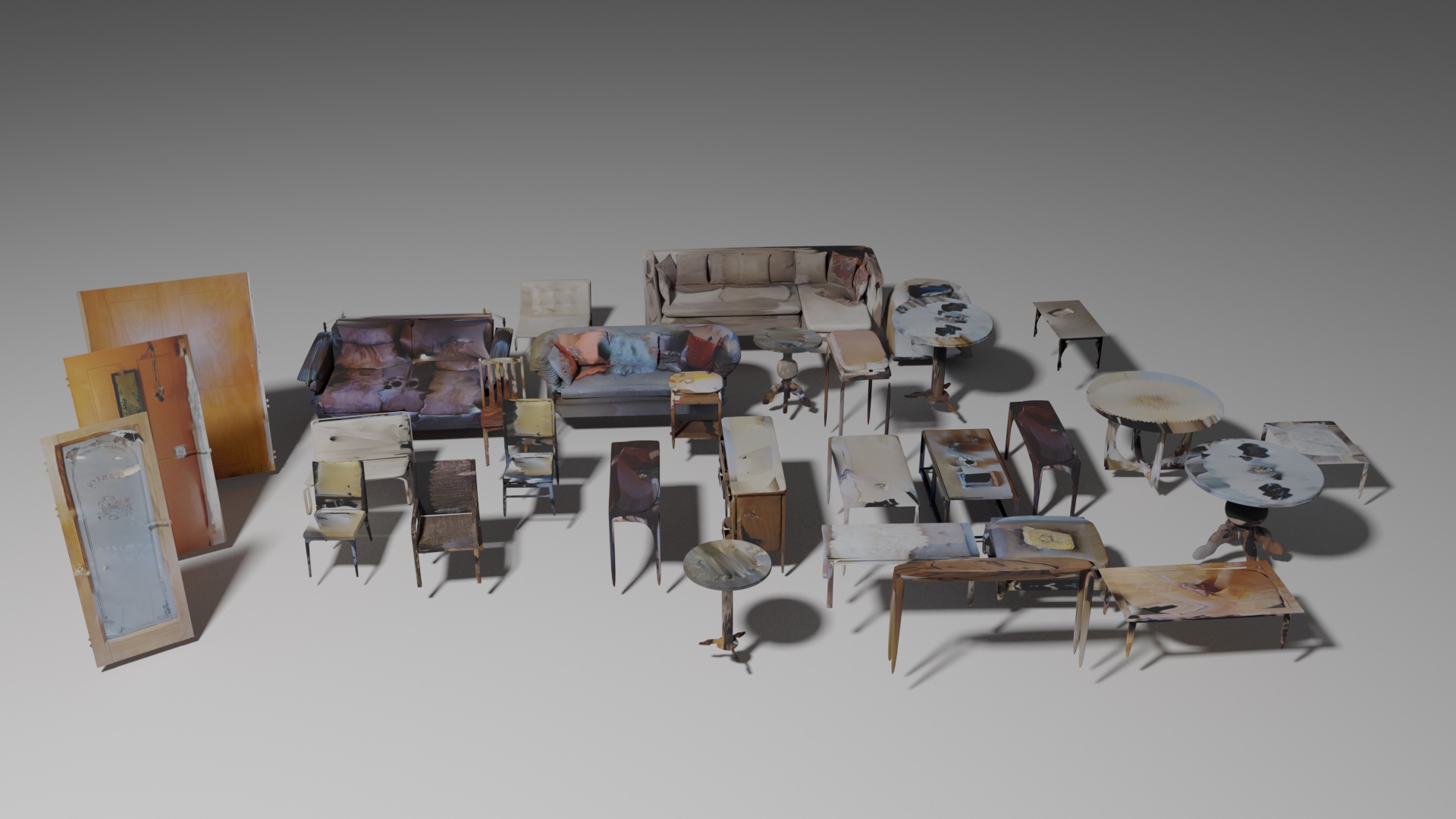}
        \caption{}
   \end{subfigure} 
~
  \begin{subfigure}{0.4\columnwidth}
         \centering
        \includegraphics[width=\textwidth]{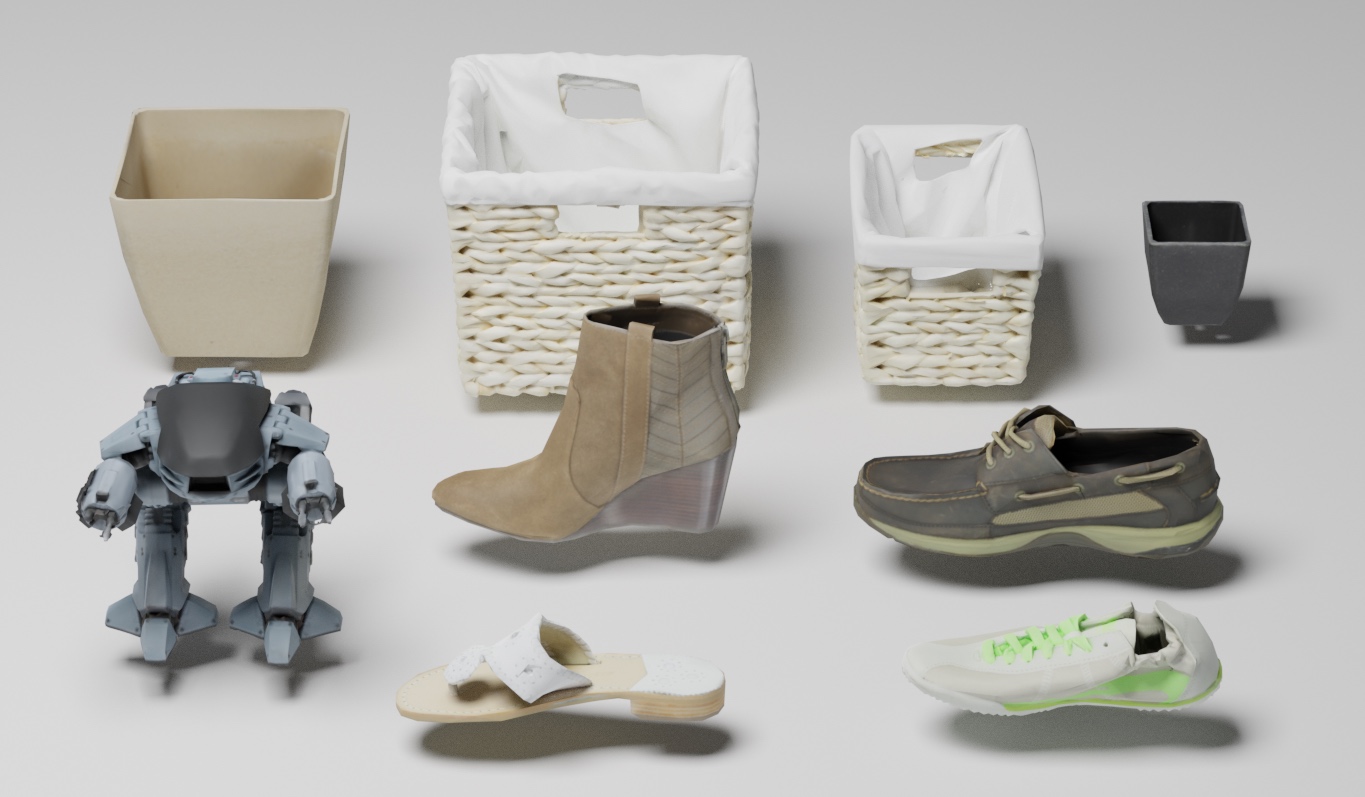}
        \caption{}
        \label{fig:added_objects:addobs}
   \end{subfigure} 
   
  \end{center}
  \vspace{-2mm}
  	\caption{\footnotesize \textbf{Interactable Objects in Interactive Gibson.} (a) Topdown view of ten 3D reconstructed scenes with objects annotated and replaced by high resolution CAD models highlighted in green. (b) Retextured  ShapeNet~\cite{chang2015shapenet} models obtained from our assisted annotation process (Sec.~\ref{ss:iga}). (c) Additional common objects randomly placed in the environment.}
	\label{fig:added_objects}
	\vspace{-2mm}
\end{figure}

As a result of our annotation and refinement of Interactive Gibson assets, the environments include interactable \revision{objects in place of} the original model parts for the following five categories: chairs, desks, doors, sofas, and tables. In addition to these existing objects in the scenes with their original poses, we add ten additional objects that are frequently found in human environments. The objects we include are baskets, shoes, pots, and toys as shown in Fig.~\ref{fig:added_objects:addobs}. The models are acquired by high resolution 3D scans of common objects.
The objects have the same weights in simulation as in the real world. The objects are randomly placed on the floor to create obstacles for the agents. 

For each episode, we randomly sample an environment, the locations to place the ten additional objects, and the starting and goal location of the agent. The episode terminates when the agent either converges to the goal location or runs out of time. The agent converges to the goal location when the distance between them is below the convergence threshold, which is defined to be the same as the agent's body width. The agent has 1,000 time steps to achieve its goal (equal to \SI{100}{\second} of wall time as if trained in the real-world). 

\subsection*{Interactive Navigation Score}

To measure Interactive Navigation performance, we propose a novel metric that captures the following two aspects:


\textbf{Path Efficiency}: how efficient the path taken by the agent is to achieve its goal. The most efficient path is the shortest path assuming no interactable obstacles are in the way. A path is considered to have zero efficiency if \revision{the agent does not converge to the goal at all.}

\textbf{Effort Efficiency}: how efficient the agent spends its effort to achieve its goal. The most efficient way is to achieve the goal without disturbing the environment or interacting with the objects. The total effort of the agent is positively correlated with the amount of energy spent moving its own body and/or pushing/manipulating objects out of its way. \revision{This aspect has been previously proposed to generate energy-efficient sample-based motion planners~\cite{sengupta2011energy, mohammed2014minimizing}}.

Path and Effort Efficiency are measured by scores, $P_{\underscoredw{eff}}$ and $E_{\underscoredw{eff}}$, respectively, in the interval $[0,1]$. In order to define Path and Effort scores, we assume there are $K$ movable objects in the scene indexed by $i\in\{1,\ldots,K\}$. For simplicity, we consider the robot as another object in the scene with index $i=0$. During a navigation run we denote $l_i$ as the length of the path taken by the $i^\textrm{th}$ object. We denote navigation success by a indicator function $\mathbb{1}_{\underscoredw{suc}}$  that takes value $1$ if the robot converges to the goal and $0$ otherwise.

Then the Path Efficiency Score is defined as the ratio between the ideal shortest path length $L^*$ computed \textit{without} any movable object in the environment, and the path \revision{length} of the robot, masked by the success indicator function:
$$
P_{\underscoredw{eff}} = \mathbb{1}_{\underscoredw{suc}}\frac{L^*}{l_0}
$$
The most path-efficient navigation would mean the robot takes the shortest path, $l_0 = L^*$
and thus $P_{\underscoredw{eff}}^*=1$. Please note that because $L^*$ is computed without any object in the environment $P_{\underscoredw{eff}}^*=1$ may not be achievable in practice, depending on if the sampled location of the objects that the robot cannot move away intersect the shortest path. This definition of Path Efficiency is equivalent to the recent metric \textit{Success
weighted by Path Length (SPL)}~\cite{anderson2018evaluation} for the evaluation of pure navigation agents.

To define the Effort Efficiency Score, we denote by $m_i$ the masses of the robot ($i=0$) and the objects. Further, $G=m_0g$ stands for the gravity force on the robot and $F_t$ stands for the amount of force applied by the robot on the environment at time $t\in[0,T]$, excluding the forces applied to the floor for locomotion. 
The Effort Efficiency Score captures both the excess of displaced mass (kinematic effort) and applied force (dynamic effort) for interactions:
$$
E_{\underscoredw{eff}}=0.5 \left(\frac{m_0l_0}{\sum_{i=0}^Km_il_i} + \frac{TG}{TG+\sum_{t=0}^TF_t}\right) 
$$
The most effort-efficient navigation policy is to not perturb any object except the robot ``itself'': $l_i=0$ for $i\in\{1, \ldots, K\}$ and $F_t=0$ for $t\in[0,T]$. In this case, $E_{\underscoredw{eff}}^*=1$.

The final metric, called \textit{Interactive Navigation Score} or \textit{INS}, captures both aspects aforementioned in a soft manner by a convex combination of Path and Effort Efficiency Scores:
\[
\textrm{INS}_{\alpha} = \alpha P_{\underscoredw{eff}} + (1-\alpha)E_{\underscoredw{eff}}
\]
\textit{INS} captures the tension between taking a short path and minimizing effort -- the robot can potentially take the shortest path (high Path Efficiency score) while pushing as many objects as needed (low Effort Efficiency score); or the robot can try to minimize effort (high Effort Efficiency score) by going around all objects and taking a longer path (low Path Efficiency score). In the evaluation we control the importance of the above trade-off by varying $\alpha\in[0,1]$, where $\alpha=1$ corresponds to the classical pure navigation SPL score\cite{anderson2018evaluation}.





\begin{figure}[t]
  \begin{center}
        \includegraphics[width=0.85\columnwidth]{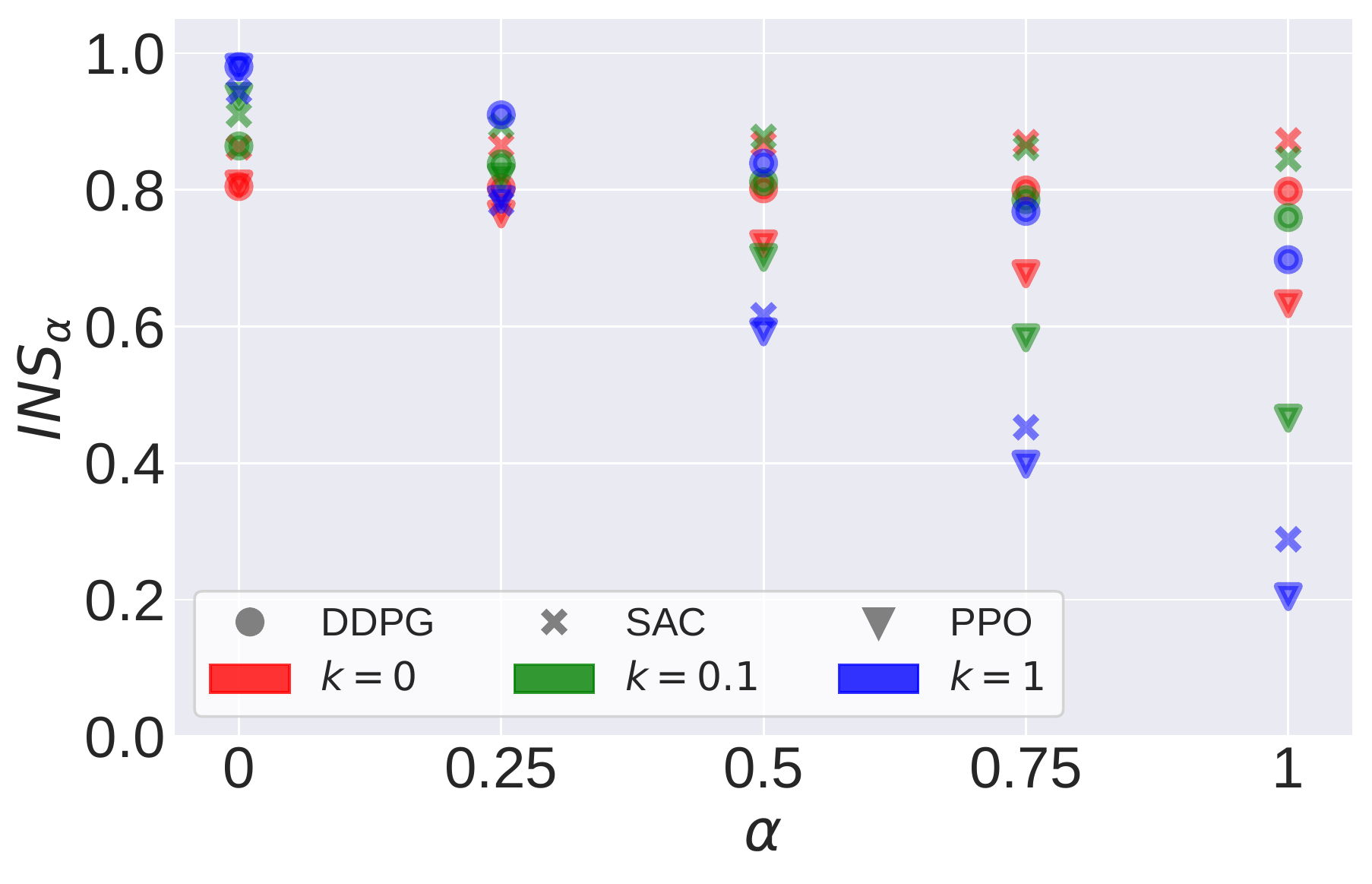}
  \end{center}
	\vspace{-3mm}
  \caption{\footnotesize \textbf{Interactive Navigation Score ($\textrm{INS}$) at different $\alpha$ levels for Turtlebot.} With $\alpha=0$ (score based only on Effort Efficiency), the best performing agents are those that minimize interactions (blue). For $\alpha=1$ (score based only on Path Efficiency, $\textrm{INS}_{1}=\textrm{SPL}$) some of these agents are overly conservative and fail to achieve the goal at all (lower $\textrm{INS}$). One of the best performing agent (SAC with $k_{int}=0.1$) strikes a healthy balance between navigation and interaction. With $\alpha=0.5$, SAC has the best performance overall except when the interaction penalty is too large ($k_{int}=1$). Markers indicate the mean of three random seeds per algorithm and interaction penalty coefficient evaluated in the two test environments.}
	\label{fig:ins}		
	\vspace{-4mm}
\end{figure}

\begin{figure}[t]
  \begin{center}
        \includegraphics[width=0.85\columnwidth]{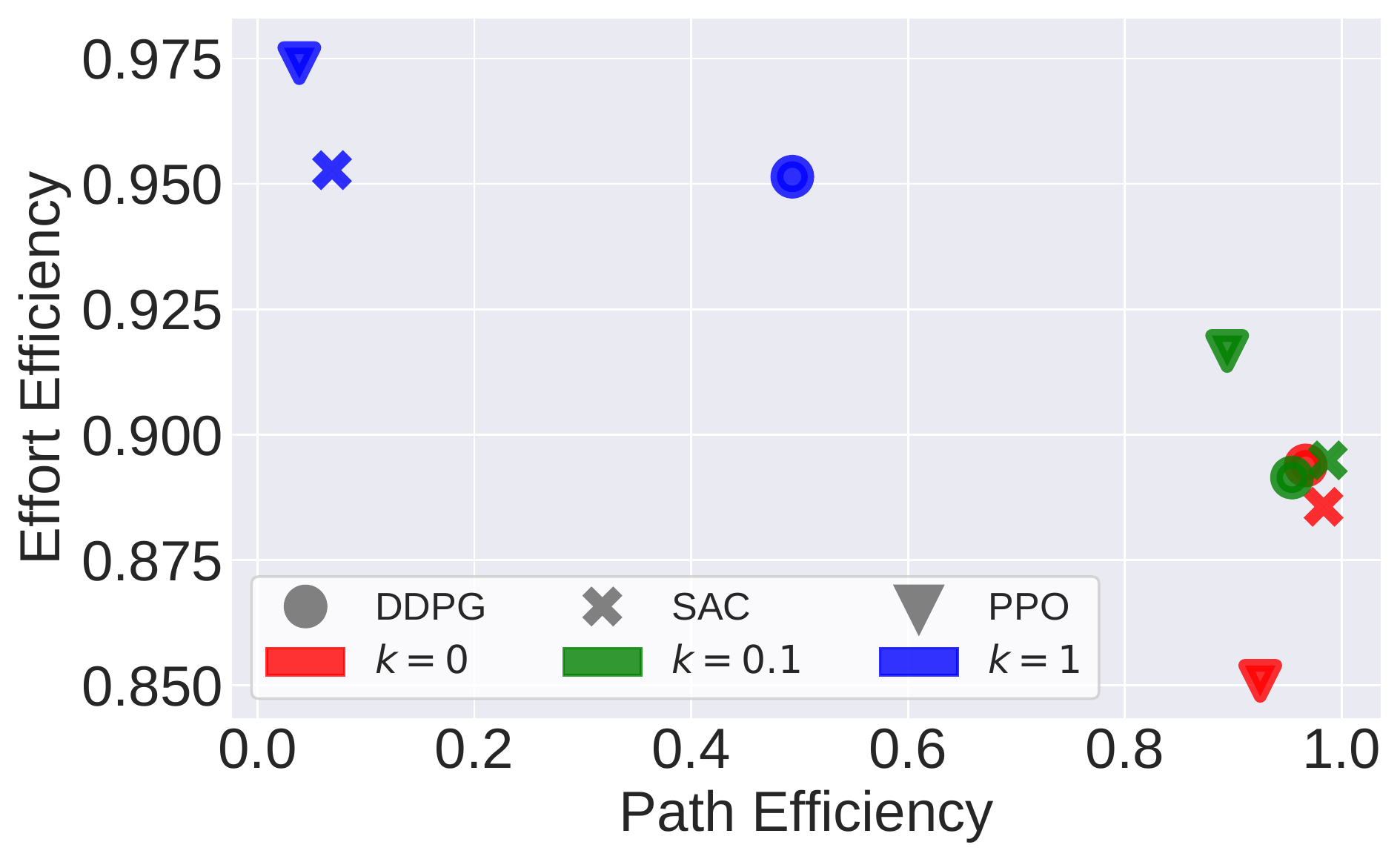}
  \end{center}
       	\vspace{-3mm}
  	\caption{\footnotesize \textbf{Trade-off between Path and Effort Efficiency for Fetch.} With high interaction penalty ($k_{int}=1$), the agents achieve higher Effort Efficiency, but at the cost of a much lower Path Efficiency. With low interaction penalty ($k_{int}=0.1$), the agents achieve almost identical Path Efficiency as those trained with no interaction penalty ($k_{int}=0$) and higher Effort Efficiency (e.g. avoiding unnecessary interactions). Markers indicate the mean of three random seeds per algorithm and interaction penalty coefficient evaluated in the two test environments.}
	\label{fig:spl_dstb}
	\vspace{-4mm}

\end{figure}

\section{Evaluating Baselines on Interactive Gibson}
\label{s:basalg}

Our goal for is to find a unified solution that can be controlled to balance path efficiency and effort efficiency. 
Although numerous learning and non-learning-based methods could potentially achieve this goal and solve Interactive Navigation, we decide, based on their versatility and recent success, to develop baselines based on well-established reinforcement learning algorithms.
We evaluate and compare three widely used reinforcement learning algorithms: PPO~\cite{schulman2017proximal}, DDPG~\cite{lillicrap2015continuous}, SAC~\cite{haarnoja2018soft} (implementations adopted from tf-agents~\cite{TFAgents} and modified to accommodate our environments). We randomly select eight Gibson scenes as our training environments and test our baseline agents in these (seen) environments and in two other (unseen) environments from the Interactive Gibson assets (scenes shown in Fig.~\ref{fig:added_objects:envs}). 

To train and evaluate our baseline agents we use two robotic platforms in simulation: TurtleBot v2 and Fetch. \revision{While Interactive Gibson Environment is able to simulate more complicated and dexterous manipulations~\cite{li2019hrl4in}, for the baselines in this work, we train agents that mainly use the body of TurtleBot v2 and Fetch to push objects}. Due to their significantly different sizes and weights, these robots interact with the objects in the environment differently. 

From the set of available observations in the Interactive Gibson Environment we employ in our baseline solutions the following: 1) goal location, 2) angular and linear velocity, and 3) the next ten waypoints of the pre-computed ground-truth shortest path, all in agent's local frame. The observation vector also includes a depth map and a semantic segmentation mask of reduced resolution ($68\times 68$). The action space for our baselines is the joint velocity of the wheels.


\textbf{Reward Function:} Our hypothesis is that the balance between path efficiency and effort efficiency (amount of interaction with the objects in the environment) can be controlled through the reward received by the RL agents. With this goal in mind we propose the following reward function:
$$
R = R_{suc} + R_{pot} + R_{int}
$$
$R_{suc}$ (\textit{suc} from \textit{success}) is a one-time sparse reward of value $10$ that the agent receives if it succeeds in the navigation (i.e. converges to the goal). $R_{pot}$ (\textit{pot} from \textit{potential}) is the difference in geodesic distance between the agent and the goal in current and previous time steps, $R_{pot}= \textit{GD}_{t-1} - \textit{GD}_{t}$ ($R_{pot}$ is positive when the distance between the agent and the goal decreases and negative when the distance increases). $R_{int}$ (\textit{int} from \textit{interaction}) is the penalty for interacting with the environment: $R_{int} = - k_{int} \mathbb{1}_{int}$. $\mathbb{1}_{int}$ is an indicator function for interaction with objects, and $k_{int}$ is the interaction penalty coefficient (positive), a hyper-parameter that represents how much the agent is penalized for interactions.

We experiment with a combination of three different interaction penalty coefficients $k_{int} = \{0, 0.1, 1.0\}$. We aim to investigate how different algorithms, robots, and (controllable) reward functions affect the navigation behavior in cluttered environments using our novel Interactive Gibson Benchmark. We train the agents in 8 environments and report the test results on 2 unseen environments. The split can be found on the project website$^3$. 



\begin{figure}[t]
  \begin{center}
    \includegraphics[width=0.92\columnwidth]{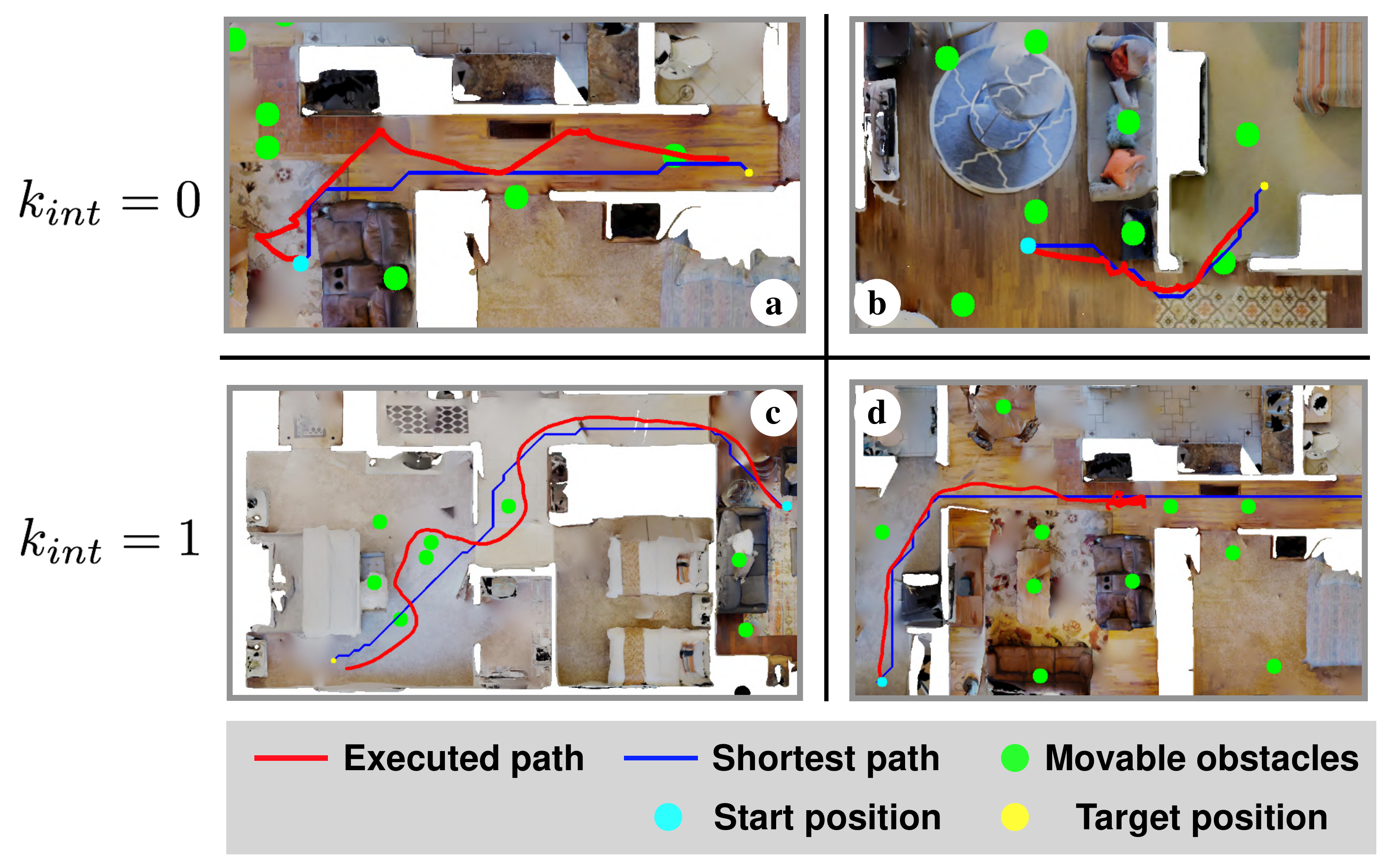}
  \end{center}
    \vspace{-2mm}
  	\caption{\footnotesize \textbf{Qualitative results of the trade-off between Path and Effort Efficiency.
  	} With no interaction penalty ($k_{int} = 0$, first row), the agent follows the shortest path computed without movable objects, and interact with the objects in its way. With high interaction penalty ($k_{int} = 1$, second row) the agent avoids collisions and deviates from the shortest path (c). It sometimes fails to achieve the goal at all when being blocked (d).}
	\label{fig:examples}
	\vspace{-8mm}
\end{figure}

\textbf{Evaluation of Baselines:} Fig.~\ref{fig:ins} depicts the Interactive Navigation Score, $\textrm{INS}_{\alpha}$, for the evaluated agents using the TurtleBot embodiment. Overall, SAC obtains the best scores independently of the relative weight between Path and Effort Efficiency Scores. Based only on the Effort Efficiency ($\alpha=0$), the best performing solutions are the ones trained to reduce interactions ($k_{int}=1$, blue). Interestingly, SAC trained to moderately reduce interactions ($k_{int}=0.1$, green) is the best performing agent independently of the balance between Path and Effort Efficiency except for $\alpha=0$. Fetch results (in our project page) present the same distribution.

Fig.~\ref{fig:spl_dstb} shows the trade-off between Path and Effort Efficiency for the agents using the Fetch embodiment. As expected, agents penalized for interacting ($k_{int}=1$, blue) obtain the best Effort Efficiency Scores at the cost of a large Path Efficiency loss: reducing interactions causes these agents to deviate more from the shortest path and even to completely fail in the navigation task (Fig.~\ref{fig:examples}, bottom row).

\begin{figure}[t]
  \begin{center}
    \includegraphics[width=0.75\columnwidth]{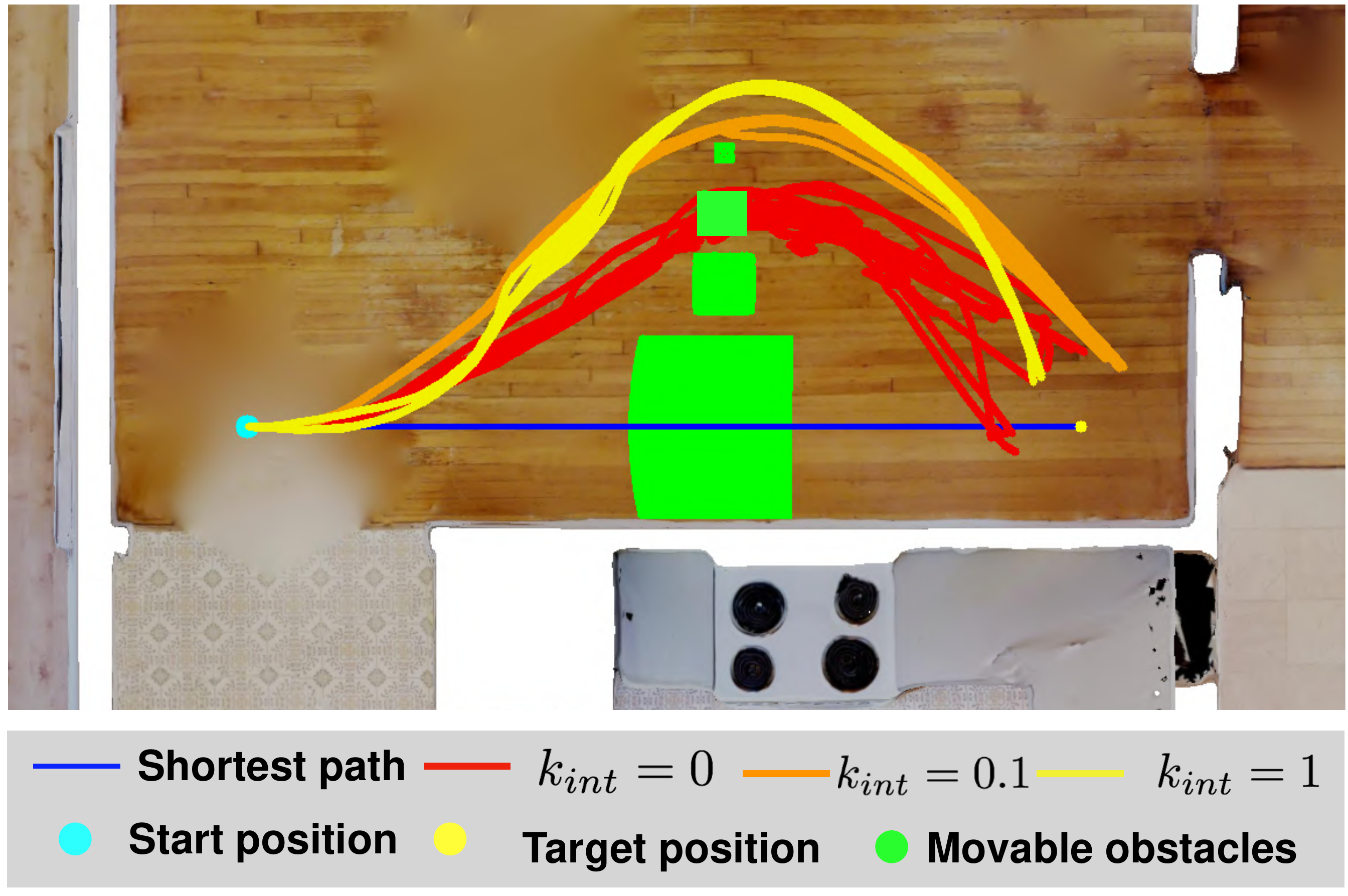}
  \end{center}
  \vspace{-2mm}
  	\caption{\footnotesize \textbf{Navigation behaviors of different interaction penalties} Top-down view of the trajectories generated by agents trained with DDPG using different interaction penalties and the TurtleBot embodiment. Depending on the penalty, the agent learns to deviate from the optimal path (blue) to avoid collisions with large objects (sofas) ($k_{int} = 0$), medium ones (baskets) ($k_{int} = 0.1$), or small ones (cups) ($k_{int} = 1$). The object class information is encoded in the semantic segmentation mask.
  	}
	\label{fig:paths}
	\vspace{-6mm}
\end{figure}

Fig.~\ref{fig:paths} shows the difference in navigation strategy of TurtleBots trained with different interaction penalties. When the penalty is high ($k_{int}=1$, yellow), the agents avoid any contact with the environment at the cost of less efficient path execution. When the interactions are less ($k_{int}=0.1$, orange) or not ($k_{int}=0$, red) penalized, the agents sacrifice effort efficiency to increase path efficiency by interacting with movable objects. Note that even without interaction penalty, the agents learn to avoid very large objects (e.g. sofas, tables) since they cannot be pushed away by TurtleBots. The agents learn this object class-specific behavior from the semantic segmentation mask generated by the Interactive Gibson Environment. Larger and more powerful Fetch robot embodied agents can also move small tables and sofas and therefore learn to interact more.

Our baselines generalize well to unseen environments: the difference in performance between the seen and unseen environments is not statistically significant. We perform a one sample t-test for evaluation results on training and test scenes measured by $\textrm{INS}_{0.5}$. The p-value is $0.171$ showing the interactive navigation solutions work equally well on unseen environments. We believe this is because, even for environments not seen during training, the robot has indirect access to the map of the environment via the shortest path input given as part of its observation (Sec.~\ref{ss:igags}). Additionally, even though the environments are different, the movable objects are of the same classes, which allows the robot to generalize how to interact or avoid collisions with them. 


\section{Conclusion and Future Work}

We presented Interactive Gibson Benchmark, a novel benchmark for training and evaluating Interactive Navigation agents.
We developed a new photo-realistic simulation environment, the Interactive Gibson Environment, that includes a new renderer and over one hundred 3D reconstructed real-world environments where all instances of object classes of relevance have been annotated and replaced by high resolution CAD models.
We also proposed \revision{a metric} called Interactive Navigation Score ($\textrm{INS}$) to evaluate Interactive Navigation solutions. INS reflects the trade-off between path and effort efficiency. 
We plan to continue annotating other object classes to extend our benchmark to other types of interactive tasks such as interactive search and retrieval of objects.
Interactive Gibson is publicly available for other researchers to test and compare their solutions for Interactive Navigation in equal conditions.

\section*{Acknowledgement}
\begingroup
\small
{The authors would like to thank Junyoung Gwak and Lyne P. Tchapmi for helpful discussions. We thank Google for funding. Fei Xia would like to thank Stanford Graduate Fellowship for the support.}
\endgroup

\bibliographystyle{IEEEtranN}
\footnotesize
\bibliography{references}

\end{document}